\documentclass{article}

\usepackage[english]{babel}
\usepackage{multirow,multicol,ctable,tabularx}
\usepackage[group-separator={,}]{siunitx}
\usepackage[cmex10]{amsmath}
\usepackage{ragged2e}
\usepackage{caption}
\usepackage{multicol}
\usepackage{balance}
\usepackage[normalem]{ulem}

\newcommand{\tname}{deepMiRGene}

\newcommand{\varA}[1]{{\operatorname{#1}}}

\usepackage{times}
\usepackage{graphicx} 

\usepackage{natbib}

\usepackage{algorithm}
\usepackage[noend]{algorithmic}

\usepackage{hyperref}

\usepackage[accepted]{mystyle}

\icmltitlerunning{}

\begin{document}

\twocolumn[
\icmltitle{\tname: Deep Neural Network based Precursor microRNA Prediction \\
           }

\icmlauthor{Seunghyun Park, Seonwoo Min, Hyun-soo Choi, and Sungroh Yoon* }{sryoon@snu.ac.kr}
\icmladdress{Department of Electrical and Computer Engineering, Seoul National University, Seoul 151-744, Korea}


\vskip 0.3in
]

\begin{abstract}

Since microRNAs (miRNAs) play a crucial role in post-transcriptional gene regulation, miRNA identification is one of the most essential problems in computational biology. miRNAs are usually short in length ranging between 20 and 23 base pairs. It is thus often difficult to distinguish miRNA-encoding sequences from other non-coding RNAs and pseudo miRNAs that have a similar length, and most previous studies have recommended using precursor miRNAs instead of mature miRNAs for robust detection. A great number of conventional machine-learning-based classification methods have been proposed, but they often have the serious disadvantage of requiring manual feature engineering, and their performance is limited as well. In this paper, we propose a novel miRNA precursor prediction algorithm, \tname, based on recurrent neural networks, specifically long short-term memory networks. \tname~ automatically learns suitable features from the data themselves without manual feature engineering and constructs a model that can successfully reflect structural characteristics of precursor miRNAs. For the performance evaluation of our approach, we have employed several widely used evaluation metrics on three recent benchmark datasets and verified that \tname~delivered comparable performance among the current state-of-the-art tools.

\end{abstract}

\section{Introduction}\label{submission}

A miRNA (microRNA) is a small non-coding RNA that plays a crucial role in post-transcriptional gene regulation by attaching itself to the 3' untranslated region of the target mRNA~\cite{lee1993c}. There are a number of research problems related to miRNA, including the search for miRNA itself or the miRNA regulation target, messenger RNA (mRNA). Among the many problems, how to computationally identify miRNAs has been one of the most significant problems. From the engineering point of view, miRNA identification can be understood as a binary classification problem that classifies input sequences into miRNA or non-miRNA. miRNA follows the sequence of primary miRNA into precursor mRNA (pre-miRNA), then into mature miRNA and RNA-induced silencing complex~\cite{bartel2004micrornas}. Mature miRNAs are usually short, having 20 to 23 base pairs (bp), and thus it is difficult to identify them using only sequence patterns. In order to identify miRNAs, most computational approaches thus focus on detecting pre-miRNAs since they are usually longer (approximately 80bp) and also have the distinguishing feature of a stem-loop secondary structure. The advent of next generation sequencing has made it possible to detect RNAs even in low concentrations. However, it has also led to the discovery of many other novel RNAs besides miRNA, such as siRNA, piRNA, and degradation products of ribosomal RNA and transfer RNA, leading to an increase in identification subjects and consequently raising the problem of high false positives~\cite{kang2015computational}. 

Many computational approaches to identifying miRNA have been proposed and can be divided into two categories: conservation and rule-based methodologies and machine-learning-based methodologies~\cite{kleftogiannis2013we}. Since a sufficient number of miRNAs for machine learning are now available, currently utilized tools are mostly machine-learning-based. Specific machine learning algorithms used are diverse. MiPred~\cite{jiang2007mipred}, microPred~\cite{batuwita2009micropred}, triplet-SVM~\cite{xue2005classification}, and miRBoost~\cite{tempel2015mirboost} uses the support vector machine (SVM); CSHMM~\cite{agarwal2010prediction} has adopted the hidden Markov model (HMM) and additionally utilized context-sensitive characteristics to consider secondary structures more carefully; and MIReNA~\cite{mathelier2010mirena} uses five rule-based schemes.

What the mentioned approaches have in common is that they use hand-crafted features that include structural and folding energy information of miRNA precursors. For example, the frequency of triplets appearing in the loop, the stem length, and minimum free energy are widely used features. Some studies have even argued that the performance of machine learning-based tools is more dependent on input feature sets rather than the specific machine-learning algorithms~\cite{de2014discriminant}. Therefore, most previous approaches have focused on either searching for novel features or combining the existing features using ensemble algorithms. Indeed, high accuracy was reported for miRBoost and microPred using more than 100 known features. Nonetheless, most of the existing tools still suffer from the low-sensitivity issue.

In this paper, we propose \tname~, which uses recurrent neural networks (RNNs), specifically long short-term memory (LSTM) networks, to learn sequence patterns and folding structure. {\bf The most important contribution of the proposed approach is that it does not require any painful manual feature engineering.} This method takes advantage of end-to-end deep learning, which only requires simple preprocessing instead of a considerable amount of domain knowledge to design hand-crafted features. Since miRNA has a palindromic secondary structure, it is difficult to immediately apply an LSTM network. To solve such difficulties, {\bf we propose a novel method for learning the palindromic secondary structure of precursor miRNA.} Furthermore, {\bf \tname~ delivers superior performance, outperforming all compared alternatives in terms of sensitivity and specificity on the benchmarking datasets.} \tname~ also gives the best performance in cross-species data, even though many differences exist between the features among the different species. Our approach shows the possibility of rediscovering intrinsic features in a data-driven fashion and is expected to bring novel biological knowledge as an automated and effective feature extractor.

\section{Related Work}
\subsection{RNN and LSTM}\label{RW:LSTM}

\begin{figure}[!tpb]
\begin{center}
\centerline{\includegraphics[width=\columnwidth]{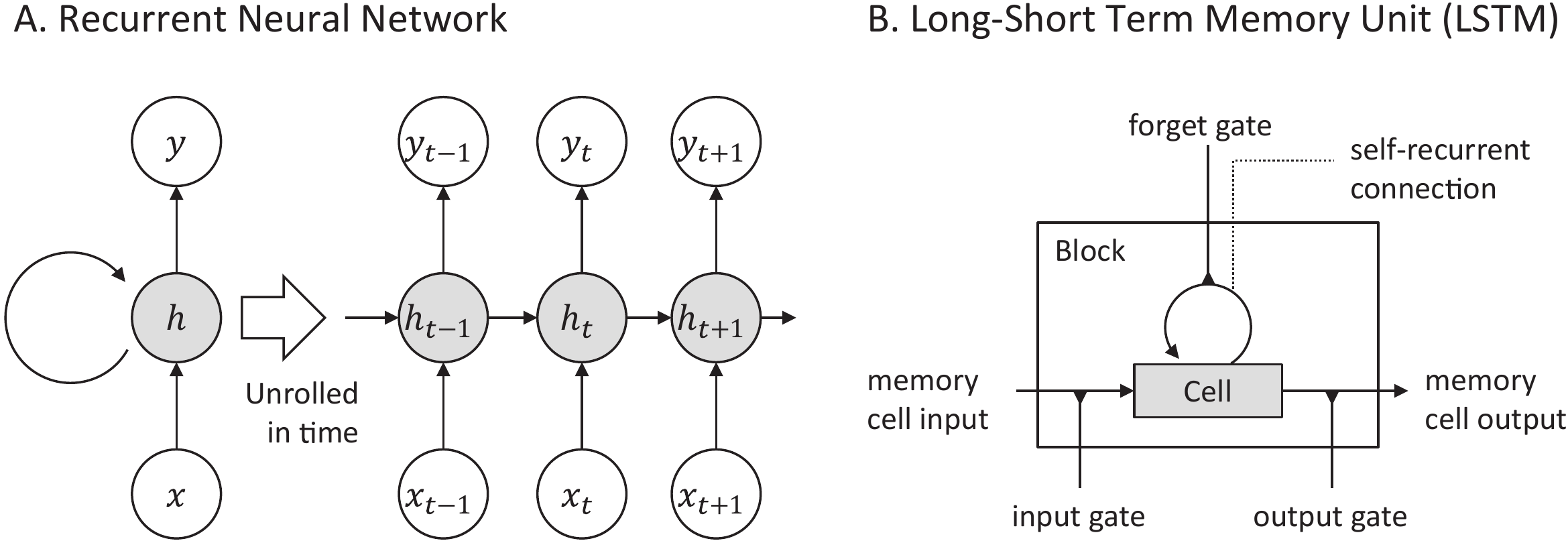}}
\caption{(A) Basic structure of recurrent neural network~\cite{lecun2015deep} and (B) diagram of LSTM cell unit~(deeplearning.net).}
\label{Fig:rnn}
\end{center}
\end{figure}

RNN is a deep learning structure designed to learn variable length sequential data. Figure~\ref{Fig:rnn}(A)~shows the basic structure of RNN. The core aspect of RNN is that unlike other structures, RNN processes input data one element at a time and stores past information implicitly using cyclic connections of hidden units~\cite{lecun2015deep}. Since time-unfolded RNN is an even deeper structure than DNN or CNN, it is difficult to learn long-term dependency with simple perceptron hidden units due to the gradient vanishing problem~\cite{bengio1994learning}. Therefore most RNN research uses more sophisticated hidden units that operate as some kind of memory cell. LSTM~\cite{hochreiter1997long}, shown in Figure~\ref{Fig:rnn}(B), is the most well-known example. Besides cyclic connections storing the “state vector,” LSTM uses multiplicative gates to learn when to input, output, and forget to produce better performance with RNN.

\setlength{\textfloatsep}{0.1cm}

\subsection{Palindromic Structure of Folded miRNA Precursor}\label{palindrome}

\begin{figure}[!tpb]
\begin{center}
\centerline{\includegraphics[width=\columnwidth]{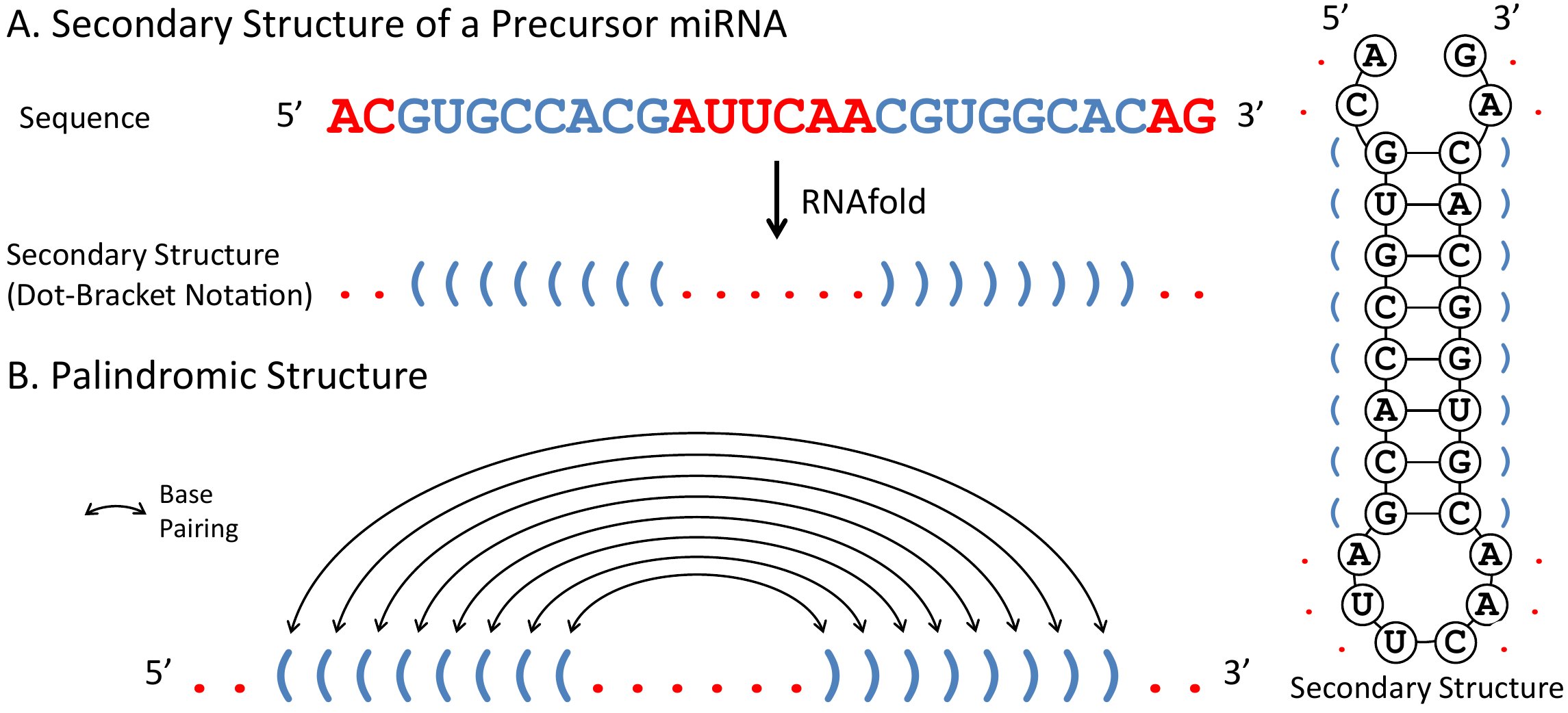}}
\caption{The secondary structure of a precursor miRNA and its panlindrome. (A) The left means the sequence of a precursor miRNA and the right represents the secondary structure of a given sequence. Dot-bracket notation (DBN), below the sequence, is the method for describing a secondary structure. Unpaired nucleotides are represented as ``.'' and base-paired nuclotides are represented as opening ``(''s and closing ``)''s. (B) A Palindrome in secondary structure. The forward strand (5'$\,\to\,$3') on the left side of the middle point and the backward strand (3'$\,\to\,$5') on the right side of the middel point match complementarily.}
\label{Fig:Fold}
\end{center}
\end{figure}

Precursor miRNA exists in the form of a base-paired double helix rather than a single strand, and its structural information is important in its identification. RNAfold~\cite{hofacker2003vienna} is a widely used tool to predict the secondary structure from a sequence. It predicts the thermodynamically stable secondary structure of a given RNA sequence by calculating the minimum free energy (MFE) and the base-pairing probabilities~\cite{lorenz2011viennarna}. The ordinary secondary structure of a precursor miRNA is shown in Figure~\ref{Fig:Fold}~(A). In dot-bracket notation (DBN), one of the widely used expression methods for secondary structure, unpaired nucleotides are represented as “.” and paired nucleotides are represented as opening ``(''s and closing ``)''s. This structure consisting of helices and a loop, is called stem-loop or hairpin structure. On the other hand, pseudo miRNA precursors and other noncoding RNAs have a structure that distinguishes them from true precursor miRNAs, such as asymmetric bulges and multiple loops. Although some false positives exist due to limitations of prediction algorithms and unpredictable structures, like pseudoknots~\cite{lyngso2004complexity}, secondary structure is still one of the most essential features for identifying precursor miRNAs.

\setlength{\textfloatsep}{0.1cm}

A notable characteristic of the stem-loop structure of a precursor miRNA is that it is palindromic. As shown in Figure~\ref{Fig:Fold}(B), the left side of the stem, which is a forward strand (5'$\,\to\,$3'), and the right side of the stem, which is a backward strand (3'$\,\to\,$5'), make complementary matches, forming a helix. Therefore, from the backward strand point of view, the stem-loop structure of the precursor miRNA constitutes a form of stack. However, since general LSTM networks are designed to learn sequential data and constitute a form of queue, it requires special preprocessing, as reversing a structure, which is discussed further in Section~\ref{method}.

\section{Methodology}\label{method}
\begin{figure*}[!tpb]
\begin{center}
\centerline{\includegraphics[width=\textwidth]{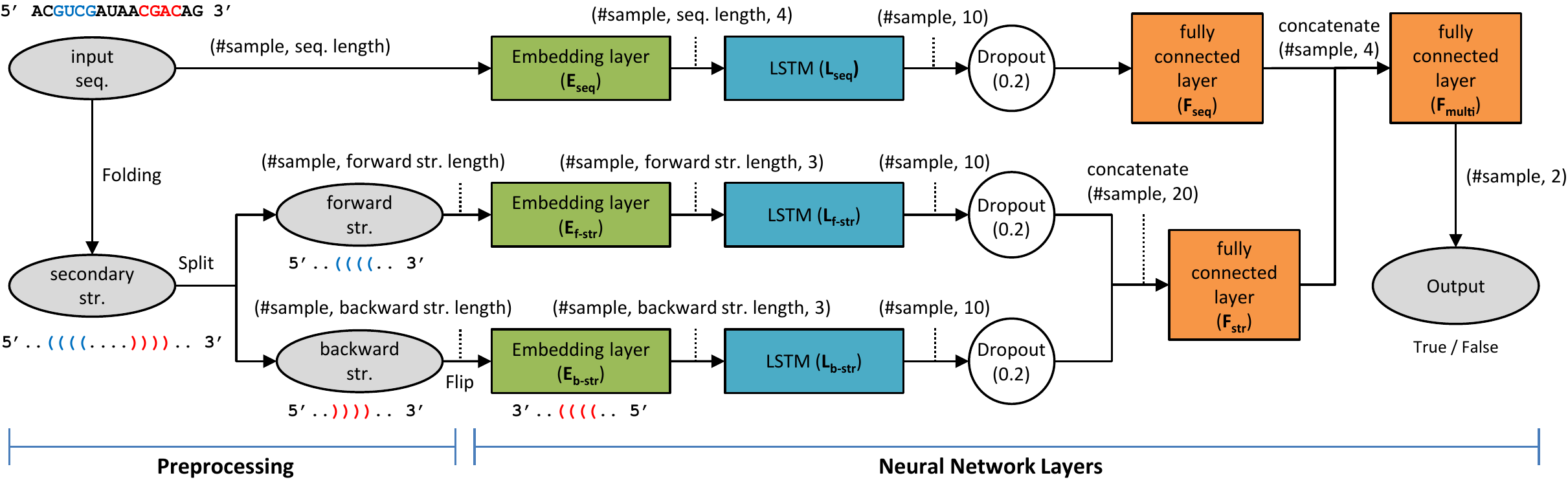}}
\caption{Overview of the proposed method.}
\label{Fig:Overview}
\end{center}
\end{figure*}
\setlength{\textfloatsep}{0.1cm}

An overview of the proposed method is shown in Figure~\ref{Fig:Overview}, and Algorithm~\ref{alg:pseudo} presents more details of the proposed approach. In the preprocessing step, the secondary structure of the input miRNA sequence is generated and split into a forward structure and a backward structure. Then in the neural network layers, RNN model parameters are trained according to the input data. The miRNA sequence, the forward structure, and the reversed backward structure are entered into the embedding layer. Finally, through three LSTM layers and fully connected layers, prediction results are produced.

\input{algorithm}

\subsection{Preprocessing}\label{preprocessing}

The secondary structure is generated using RNAfold and can be divided into forward structure, which has a direction of 5' to 3', and backward structure, which has a direction of 3' to 5'. In this study, based on the loop center of the secondary structure, the 5' side is categorized as forward and the 3' side is categorized as backward. If multiple loops exist, the middle point between the start point of the 5' closest loop and the end point of the 5' farthest loop is used as the basis.

\subsection{Construction of RNN Model}\label{nnlayer}

{\bf Embedding Layers:} miRNA sequence and structure are categorical data that have observable states of four (A, C, G, U) and three ``('', ``.'', ``)'', respectively. Thus word embedding layers are added to embed the miRNA sequence into four dimension ($\varA{E_{seq}}$) and structure into three dimension ($\varA{E_{f-str}}$ and $\varA{E_{b-str}}$). The embedding layer does not use one-hot encoding, but rather adopts weight matrices to learn the proper encoding from data as well.

{\bf LSTM Layers:} Embedded data streams are entered into three independent LSTM layers. All of the $\varA{L_{seq}}$, $\varA{L_{f-str}}$, and $\varA{L_{b-str}}$ layers have 10 hidden nodes as outputs and use hyperbolic tangent and hard sigmoid as their inner activation functions.

{\bf Fully Connected Layers:} Outputs of three LSTM layers are first connected to two fully connected layers. $\varA{F_{seq}}$ receives 10-dimension input from $\varA{L_{seq}}$, and $\varA{F_{str}}$ receives 20-dimension input from concatenation of $\varA{L_{f-str}}$ and  $\varA{L_{b-str}}$. Both $\varA{F_{seq}}$ and $\varA{F_{str}}$ have an output of two dimensions and their concatenation is connected to the final fully connected layer $\varA{F_{multi}}$, which has output of two dimensions as well. All fully connected layers use sigmoid function as their activation functions. For regularization, several methods can be used such as dropout or batch normalization~\cite{ioffe2015batch}. In this study, we selected dropout with a rate of 0.2.

Training settings are the same as follows. The mean squared error (MSE) is used as an objective function, and Adam~\cite{DBLP:journals/corr/KingmaB14} is used as the optimizer. Adam is one of the gradient descent algorithms, which computes adaptive learning rates for each parameter similar to momentum. In other words, Adam considers the moments of the gradient, such as RMSProp.

\subsection{Experimental Setup}

Three kinds of benchmark datasets from miRBoost~\cite{tempel2015mirboost} were used. The number of human, cross-species, and new pre-miRNAs datasets is shown in Table~\ref{Tab:datasets}.

The algorithm is implemented by the Theano~\cite{Bastien-Theano-2012,bergstra+al:2010-scipy} and Keras~\cite{chollet2015keras} library. The five fold cross-validations are carried out for all data, and the mini-batch size and training epoch are set as 128 and 500 times, respectively. The experiment was performed on a server consisting of an Intel Xeon E5-2650 and Nvidia Geforce Titan GPU.

\ctable[
	caption = The number of datasets used in this study,
	label = Tab:datasets,
	doinside = \footnotesize,
	width = \columnwidth,
]{lccr}{}
{\FL
Type & Human & Cross-species & New pre-miRNAs \ML
Positive set	& \num{863}	&	\num{1677} &	\num{690} 	\NN
Negative set	& \num{7422} 	&	\num{8266} &	\num{8246} 	\LL
}
\setlength{\textfloatsep}{0.5cm}

\section{Experimental Results}
\subsection{Performance Evaluation}

\ctable[
	caption = Performance evaluation.,
	label = Tab:evaluation,
	doinside = \footnotesize,
	star,
]{lcccccccccccr}{}
{\FL
	& \multicolumn{4}{c}{Human}	& \multicolumn{4}{c}{Cross-species} & \multicolumn{4}{c}{New pre-miRNAs} \ML
    Software & \textit{SE} & \textit{SP} & F-score & g-mean & \textit{SE} & \textit{SP} & F-score & g-mean & \textit{SE} & \textit{SP} & F-score & g-mean \\
    \textit{miRBoost} & 0.82  & 0.98  & 0.89  & 0.90  & 0.84  & 0.97  & 0.90  & 0.90  & \textbf{0.88}  & 0.91  & 0.89  & 0.89  \\
    \textit{CSHMM} & 0.49  & \textbf{0.99} & 0.65  & 0.70  & 0.42  & 0.97  & 0.58  & 0.64  & 0.24  & 0.95  & 0.37  & 0.48  \\
    \textit{triplet-SVM} & 0.67  & 0.98  & 0.79  & 0.81  & 0.74  & 0.96  & 0.83  & 0.84  & 0.41  & 0.95  & 0.56  & 0.62  \\
    \textit{microPred} & 0.76  & \textbf{0.99} & 0.86  & 0.87  & 0.82  & \textbf{0.98} & 0.89  & 0.90  & 0.72  & 0.97  & 0.82  & 0.84  \\
    \textit{MIReNA} & 0.83  & 0.92  & 0.87  & 0.87  & 0.80  & 0.93  & 0.86  & 0.86  & 0.46  & 0.91  & 0.59  & 0.65  \ML
    \textit{\tname} & \textbf{0.89 } & \textbf{0.99} & \textbf{0.93}  & \textbf{0.94}  & \textbf{0.91} & \textbf{0.98} & \textbf{0.94}  & \textbf{0.94}  & \textbf{0.88} & \textbf{0.99} & \textbf{0.93}  & \textbf{0.94} \LL
}

Table~\ref{Tab:evaluation} shows benchmarking results on three different datasets. Sensitivity and specificity values of all compared tools are based on experimental results of MiRBoost. Evaluation metrics, such as accuracy, positive predictive value (PPV), F-score, Matthews correlation coefficient (MCC), and the geometric mean (g-mean), are dependent on the proportion of positive and negative data in the test dataset. In this paper, we set the proportion equally and calculated the evaluation metrics.

The results show that \tname~gives the best performance in every evaluation metric. In the human dataset, all of the tools maintained high specificity, but \tname~achieved 6 percentage points higher sensitivity than MIReNA and 4 percentage points higher F-score than miRBoost, which are the highest among the conventional tools, respectively. Similarly in the cross-species dataset, the proposed method achieved 7 percentage points higher sensitivity and 4 percentage points higher F-score than miRBoost, which ranks the highest among the conventional tools. Finally in the new pre-miRNAs dataset, although \tname~shows the same level of sensitivity as miRBoost, it achieved the higher specificity by 8 percentage points.

\begin{figure}[!tpb]
\begin{center}
\centerline{\includegraphics[width=\columnwidth]{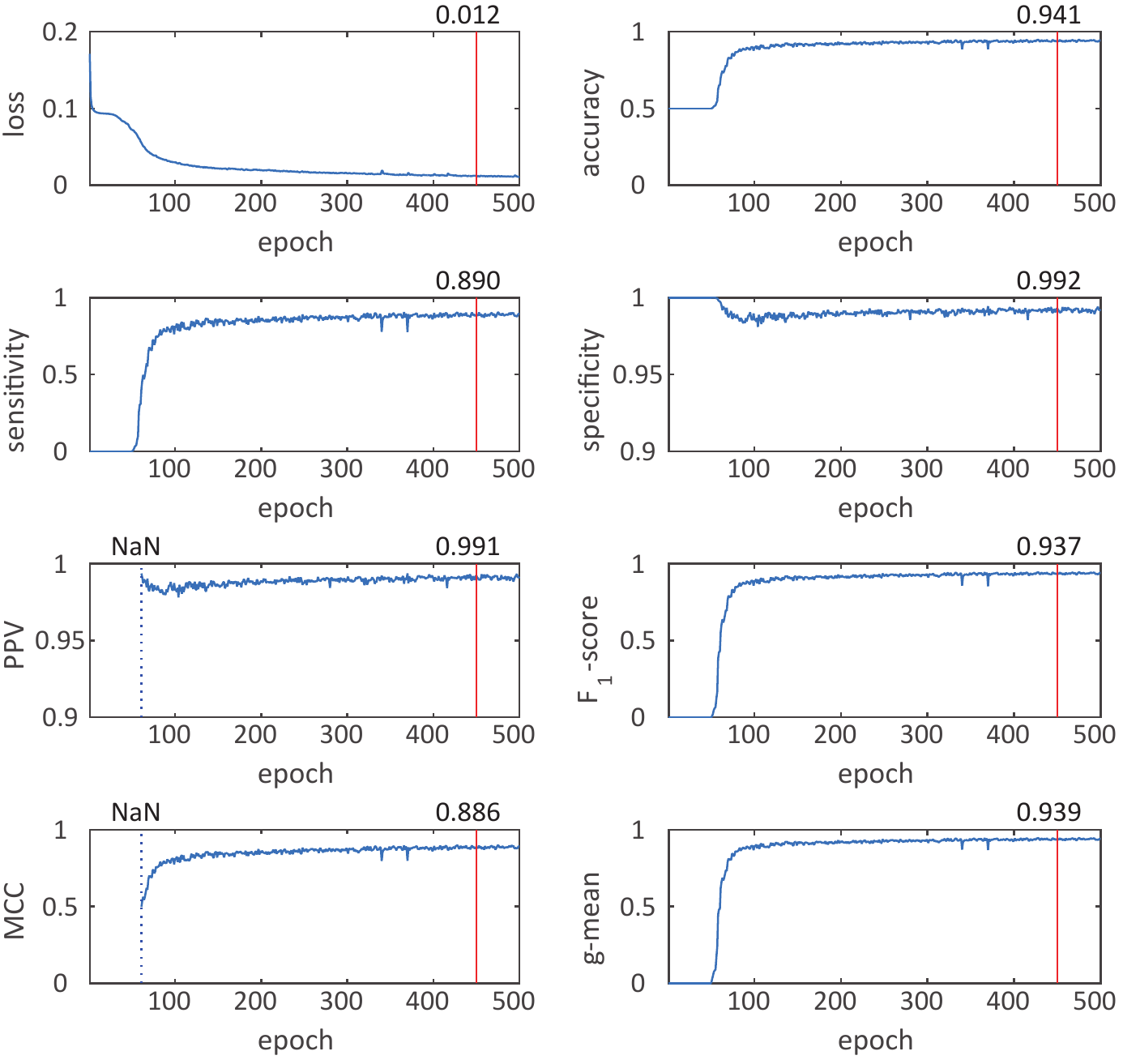}}
\caption{Training loss and seven evaluation metrics using test dataset with a varying epoch.}
\label{Fig:loss}
\end{center}
\end{figure}
\setlength{\textfloatsep}{0.1cm}

The performance evaluation metrics of \tname~are calculated as follows. A five fold cross-validation was carried out and assuming epoch 450 to 500 as the interval of convergence, we averaged the metric values in the range. Figure~\ref{Fig:loss} shows the change in training loss and evaluation metrics relative to the training epoch number, and the interval of convergence is marked in red in each graph. Parts without values indicate not a number (NaN). Specificity constantly showing a value close to 1 and other metrics showing the increase as the training epoch progresses can be understood in terms of imbalance of training data. As in Table~\ref{Tab:datasets}, negative data are relatively larger than positive data in the training dataset. Therefore, in the early training phase, prediction is biased toward the negative dataset and as learning progresses sufficiently, the prediction is tuned and converged.

\subsection{Effect of Multimodality}

\begin{figure}[!tpb]
\begin{center}
\centerline{\includegraphics[width=\columnwidth]{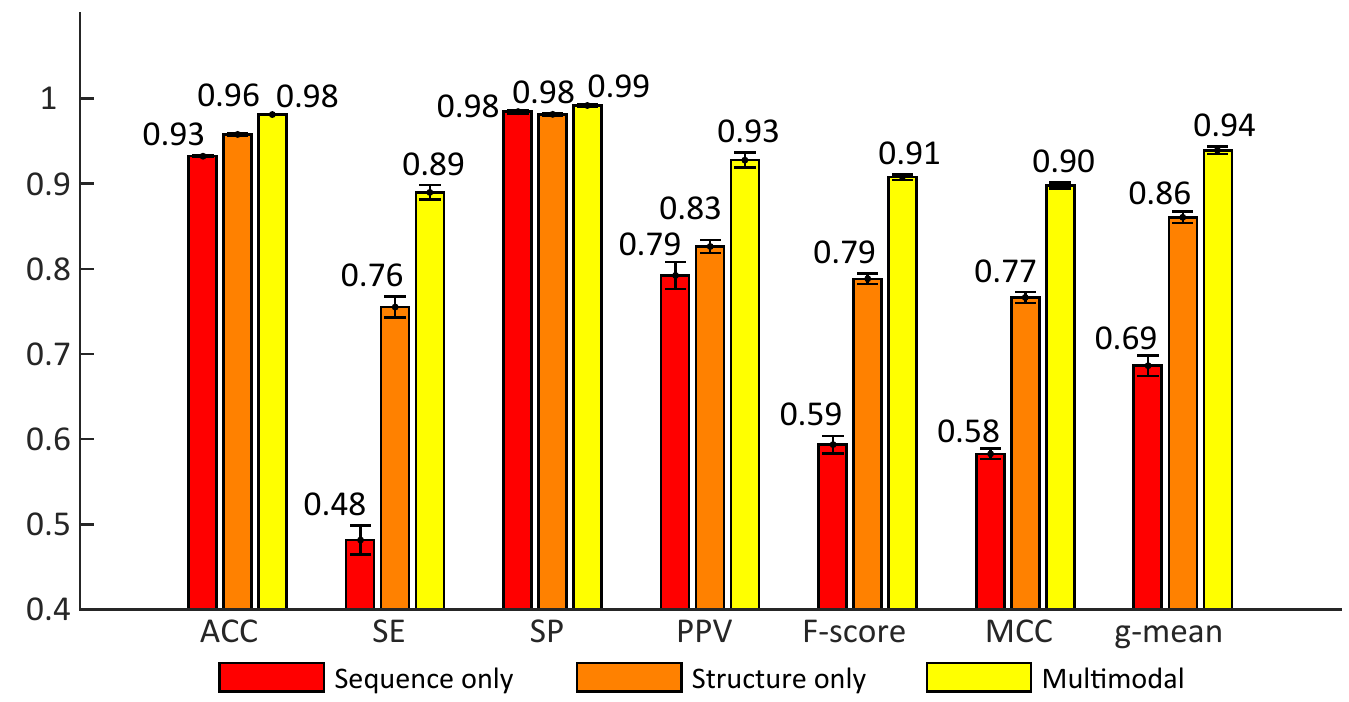}}
\caption{Performance evaluation of the human dataset in cases of using only sequence, only structure, and both sequence and structure (multimodal). Average scores of the 5 fold results are reported together with the error bars.}
\label{Fig:multimodality}
\end{center}
\end{figure}

In this study, we took advantage of both biological sequence information and derived secondary structure information in miRNA classification. To verify the effect of multimodality, we tested cases of either biological sequence or secondary structure information is utilized for the human dataset. As shown in Figure~\ref{Fig:multimodality}, all of the performance metrics are higher when both types of information are used. To be specific, multimodality achieved 41 and 13 percentage points higher sensitivity than when only sequence or structure was utilized, respectively. Similarly, in terms of F-score, multimodality showed 12 and 32 percentage points higher scores. Between sequence and structure information, derived structure information seems to have a more direct effect on accuracy than sequence information.

\subsection{Learning Capability of Palindromic Structure}

\begin{figure}[!tpb]
\begin{center}
\centerline{\includegraphics[width=\columnwidth]{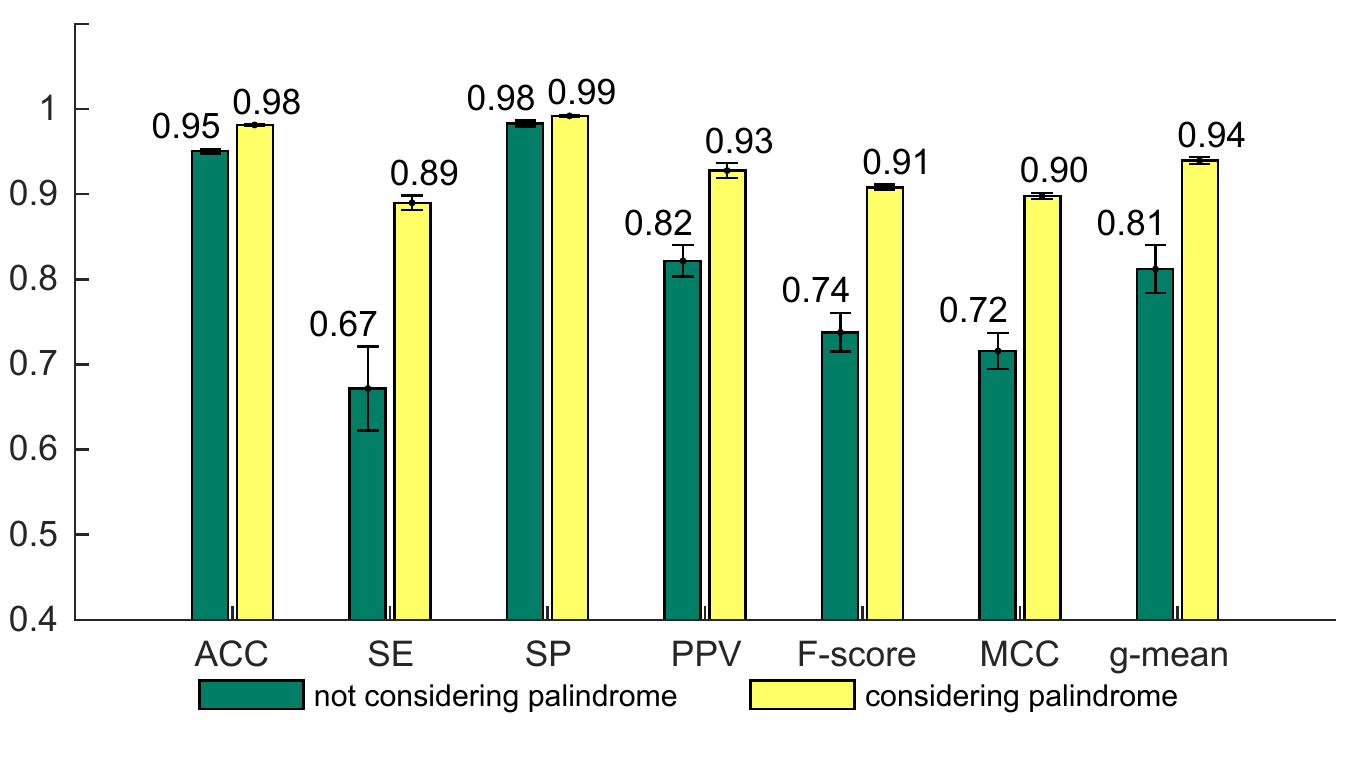}}
\caption{Performance evaluation of human dataset in the cases of considering palindromic structure or not. Average scores of the 5 fold results are reported together with the error bars.}
\label{Fig:palindrome}
\end{center}
\end{figure}

In the proposed method, structure preprocessing of split and flip was used to properly learn palindromic structures. Figure~\ref{Fig:palindrome} shows the performance comparison in the case of considering the palindromic structure or not. In all of the compared evaluation metrics, considering the palindromic structure produced the better results. Especially, the difference of specificity is up to 22 percentage points. As mentioned in Section~\ref{palindrome}, it is because the general LSTM structure is designed to learn from sequential data. Therefore, we were able to verify that the preprocessing step adopted in the proposed method can help effectively learn palindromic structures.

\subsection{Visualization of Cell State Transition}

\begin{figure*}[!t]
\begin{center}
\centerline{\includegraphics[width=\textwidth]{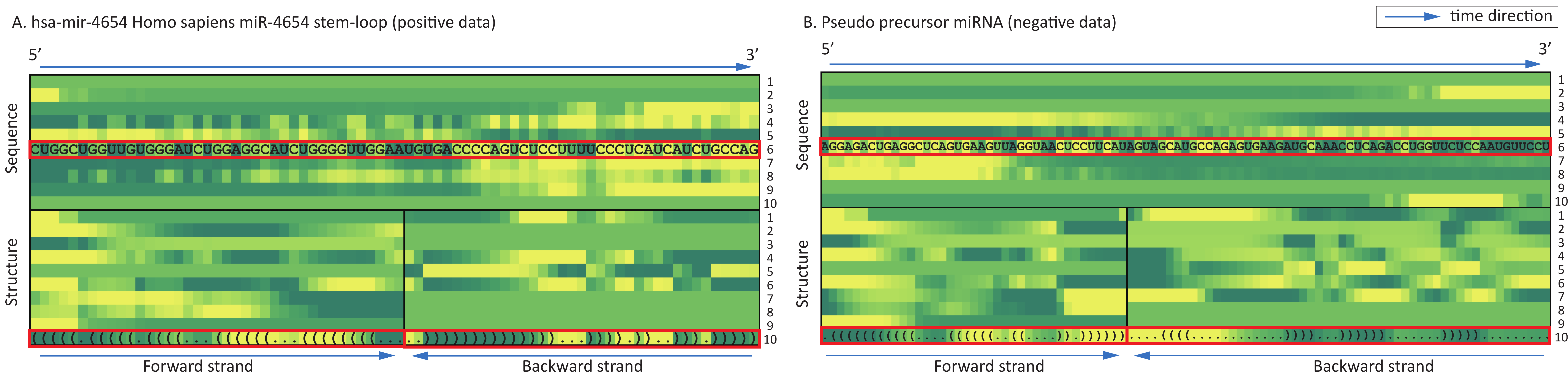}}
\caption{The transition of LSTM cell states related sequence and structural information for a positive (A) and a negative data sample (B)}
\label{Fig:intermediate}
\end{center}
\end{figure*}
\setlength{\textfloatsep}{0.5cm}

RNNs, specifically LSTM network classification methods usually contain features implicitly in the intermediate layer. There is a downside to these high-level features, in that it is difficult to intuitively understand them. Therefore much research is being conducted to find low-level features~\cite{karpathy2015visualizing,li2015visualizing}. Visualization of low-level features is dependent on the problem to be solved; thus a range of high to low level approaches are needed. Many features used in machine-learning-based methodologies, such as microPred and miRBoost, are related to the secondary structure of the precursor miRNA. For example, there is frequency of triplets in the sequence, length of stem and loop structures, folding energy, and so on.

Figure~\ref{Fig:intermediate} shows the transition of cell states while positive and negative data are being processed by the trained model. LSTM networks in this paper consist of 10 hidden nodes, so for each sequence and structure they are presented as a heatMap. The top parts of (A) and (B) show the cell states related to the sequence, and the bottom shows the cell states related to the structure. In the top red boxes, intensity differences exist between nucleotides (A,U) and (G,C). Since nucleotide pairs of A-U and G-C make hydrogen bonds that have a great influence on the structure of miRNA sequences, differences in the LSTM cell states can be understood as one of the successfully learned structural features. In the bottom red boxes of Figure~\ref{Fig:intermediate}A, most boundaries between the continuous dots (loop/bulge) and the continuous brackets (stem) are clearly distinguishable by the LSTM cell states. However, in the bottom red boxes of Figure~\ref{Fig:intermediate}B, the left side of the backward strand shows different patterns. This is because the corresponding part belongs to the additional loop that deforms the palindromic structure, so it can also be understood as another learned feature to identify negative data. The notable aspect is that hidden nodes with almost no change are observed in both sequence and structure. These nodes can be understood as uninfluential ones and can be used to decide the appropriate number of nodes.

\section{Discussion}

As mentioned above, our proposed method has the clear advantage of not requiring hand-crafted features. From the engineering point of view, producing good performance is more important than the meaning of the used features. However, from the biology point of view, the meaning of the used features is also important since they are crucial in understanding the biological mechanisms. In biology, using a ``black-box'' approach whose internals cannot be interpreted is discouraged, and the visualization of cell states and activation according to time can be helpful for avoiding such a black-box situation. In this paper, we suggested a visualization method for high-level features of sequence secondary structures. Furthermore, if intuitive low-level features can be visualized, we believe that new features can also be discovered therefrom.

Since the RNAfold tool also provides computed results of images when producing the secondary structure of input RNA, it seems natural to use them in the miRNA precursor prediction as well. In this work, although details are not covered, we have applied convolutional neural networks (CNNs) which are widely used for analyzing image data. However, we only observed accuracy degradation by using CNNs, while the training time greatly increased. Because images contain more information including those contained in the dot-bracket notation, the utilization of images will eventually be helpful for further performance improvements, albeit the negative preliminary result. For future modifications, we believe that more sophisticated preprocessing techniques and model compositions to reflect miRNA secondary structure image characteristics will be needed.

One of the most important characteristics of the deep neural network is that hyperparameters, such as the number of layers and hidden units, also have great influence on the performance. We tried a 2-layer LSTM network, pretraining with an LSTM-based autoencoder; and bidirectional LSTM networks, to name a few. However, the results from varying hyperparameters and architectures were not noticeably better compared to those reported in this paper. This study has great significance, in that LSTM networks were successfully applied to the challenging problem of precursor miRNA prediction and produced the best result among the currently existing tools. Even more thorough hyperparameter optimization for additional performance boosts will be considered in our future work.

The total time spent on a single run of training was approximately 14 hours ($\approx$ 20 second $\times$ 5 fold $\times$ 500 epoch). Although training takes some time (which is also one of the main issues in deep learning), it will not be a serious drawback in our case, since repetitive training is usually not necessary. Additionally, since the prediction time is comparable to that of the other tools once training has been done, we believe that \tname~can be an appealing solution for researchers in search of a tool with accurate and robust detection performance.

\section{Conclusion}

Conventional methods for precursor miRNA identification exploit hand crafted feature sets obtained by laborious feature engineering. Many features associated with the structural characteristics have been discovered in related research, but the performance of existing approaches measured in terms of accuracy is still limited. Worse, it is becoming more and more difficult to find new effective features manually, given that more than 100 features have already been identified.

In our study, we have proposed \tname, a novel end-to-end learning approach that can identify precursor miRNAs using the RNNs, specifically LSTM networks. The proposed method has a major advantage over existing alternatives in that no hand-crafted feature set is needed and it delivers better performance in terms of all the evaluation metrics considered. The structure of a precursor miRNA is a palindromic, which is difficult to learn even with ordinary LSTM or bidirectional LSTM networks. To address this issue, \tname~uses a novel learning scheme in which the secondary structure of the input sequence is divided into the forward and backward streams and each structure stream is learned in a different sequential direction. By applying the proposed learning method, we expect an effective learning process on the data that may have conflicts in temporal direction. In addition, we confirmed the possibility of rediscovering existing structural features by visually inspecting the transition of the LSTM cell states on each position in the sequence.



\balance

\bibliography{references}

\begin{thebibliography}{24}
\providecommand{\natexlab}[1]{#1}
\providecommand{\url}[1]{\texttt{#1}}
\expandafter\ifx\csname urlstyle\endcsname\relax
  \providecommand{\doi}[1]{doi: #1}\else
  \providecommand{\doi}{doi: \begingroup \urlstyle{rm}\Url}\fi

\bibitem[Agarwal et~al.(2010)Agarwal, Vaz, Bhattacharya, and
  Srinivasan]{agarwal2010prediction}
Agarwal, Sumeet, Vaz, Candida, Bhattacharya, Alok, and Srinivasan, Ashwin.
\newblock Prediction of novel precursor mirnas using a context-sensitive hidden
  markov model (cshmm).
\newblock \emph{BMC bioinformatics}, 11\penalty0 (Suppl 1):\penalty0 S29, 2010.

\bibitem[Bartel(2004)]{bartel2004micrornas}
Bartel, David~P.
\newblock Micrornas: genomics, biogenesis, mechanism, and function.
\newblock \emph{cell}, 116\penalty0 (2):\penalty0 281--297, 2004.

\bibitem[Bastien et~al.(2012)Bastien, Lamblin, Pascanu, Bergstra, Goodfellow,
  Bergeron, Bouchard, and Bengio]{Bastien-Theano-2012}
Bastien, Fr{\'{e}}d{\'{e}}ric, Lamblin, Pascal, Pascanu, Razvan, Bergstra,
  James, Goodfellow, Ian~J., Bergeron, Arnaud, Bouchard, Nicolas, and Bengio,
  Yoshua.
\newblock Theano: new features and speed improvements.
\newblock Deep Learning and Unsupervised Feature Learning NIPS 2012 Workshop,
  2012.

\bibitem[Batuwita \& Palade(2009)Batuwita and Palade]{batuwita2009micropred}
Batuwita, Rukshan and Palade, Vasile.
\newblock micropred: effective classification of pre-mirnas for human mirna
  gene prediction.
\newblock \emph{Bioinformatics}, 25\penalty0 (8):\penalty0 989--995, 2009.

\bibitem[Bengio et~al.(1994)Bengio, Simard, and Frasconi]{bengio1994learning}
Bengio, Yoshua, Simard, Patrice, and Frasconi, Paolo.
\newblock Learning long-term dependencies with gradient descent is difficult.
\newblock \emph{Neural Networks, IEEE Transactions on}, 5\penalty0
  (2):\penalty0 157--166, 1994.

\bibitem[Bergstra et~al.(2010)Bergstra, Breuleux, Bastien, Lamblin, Pascanu,
  Desjardins, Turian, Warde-Farley, and Bengio]{bergstra+al:2010-scipy}
Bergstra, James, Breuleux, Olivier, Bastien, Fr{\'{e}}d{\'{e}}ric, Lamblin,
  Pascal, Pascanu, Razvan, Desjardins, Guillaume, Turian, Joseph, Warde-Farley,
  David, and Bengio, Yoshua.
\newblock Theano: a {CPU} and {GPU} math expression compiler.
\newblock In \emph{Proceedings of the Python for Scientific Computing
  Conference ({SciPy})}, June 2010.
\newblock Oral Presentation.

\bibitem[Chollet(2015)]{chollet2015keras}
Chollet, Fran{\c{c}}ois.
\newblock Keras: Theano-based deep learning library.
\newblock \emph{Code: https://github. com/fchollet. Documentation:
  http://keras. io}, 2015.

\bibitem[de~ON~Lopes et~al.(2014)de~ON~Lopes, Schliep, and
  de~Carvalho]{de2014discriminant}
de~ON~Lopes, Ivani, Schliep, Alexander, and de~Carvalho, Andr{\'e} CP de~LF.
\newblock The discriminant power of rna features for pre-mirna recognition.
\newblock \emph{BMC bioinformatics}, 15\penalty0 (1):\penalty0 1, 2014.

\bibitem[Hochreiter \& Schmidhuber(1997)Hochreiter and
  Schmidhuber]{hochreiter1997long}
Hochreiter, Sepp and Schmidhuber, J{\"u}rgen.
\newblock Long short-term memory.
\newblock \emph{Neural computation}, 9\penalty0 (8):\penalty0 1735--1780, 1997.

\bibitem[Hofacker(2003)]{hofacker2003vienna}
Hofacker, Ivo~L.
\newblock Vienna rna secondary structure server.
\newblock \emph{Nucleic acids research}, 31\penalty0 (13):\penalty0 3429--3431,
  2003.

\bibitem[Ioffe \& Szegedy(2015)Ioffe and Szegedy]{ioffe2015batch}
Ioffe, Sergey and Szegedy, Christian.
\newblock Batch normalization: Accelerating deep network training by reducing
  internal covariate shift.
\newblock \emph{arXiv preprint arXiv:1502.03167}, 2015.

\bibitem[Jiang et~al.(2007)Jiang, Wu, Wang, Ma, Sun, and Lu]{jiang2007mipred}
Jiang, Peng, Wu, Haonan, Wang, Wenkai, Ma, Wei, Sun, Xiao, and Lu, Zuhong.
\newblock Mipred: classification of real and pseudo microrna precursors using
  random forest prediction model with combined features.
\newblock \emph{Nucleic acids research}, 35\penalty0 (suppl 2):\penalty0
  W339--W344, 2007.

\bibitem[Kang \& Friedl{\"a}nder(2015)Kang and
  Friedl{\"a}nder]{kang2015computational}
Kang, Wenjing and Friedl{\"a}nder, Marc~R.
\newblock Computational prediction of mirna genes from small rna sequencing
  data.
\newblock \emph{Frontiers in bioengineering and biotechnology}, 3, 2015.

\bibitem[Karpathy et~al.(2015)Karpathy, Johnson, and
  Li]{karpathy2015visualizing}
Karpathy, Andrej, Johnson, Justin, and Li, Fei-Fei.
\newblock Visualizing and understanding recurrent networks.
\newblock \emph{arXiv preprint arXiv:1506.02078}, 2015.

\bibitem[Kingma \& Ba(2014)Kingma and Ba]{DBLP:journals/corr/KingmaB14}
Kingma, Diederik~P. and Ba, Jimmy.
\newblock Adam: {A} method for stochastic optimization.
\newblock \emph{CoRR}, abs/1412.6980, 2014.
\newblock URL \url{http://arxiv.org/abs/1412.6980}.

\bibitem[Kleftogiannis et~al.(2013)Kleftogiannis, Korfiati, Theofilatos,
  Likothanassis, Tsakalidis, and Mavroudi]{kleftogiannis2013we}
Kleftogiannis, Dimitrios, Korfiati, Aigli, Theofilatos, Konstantinos,
  Likothanassis, Spiros, Tsakalidis, Athanasios, and Mavroudi, Seferina.
\newblock Where we stand, where we are moving: Surveying computational
  techniques for identifying mirna genes and uncovering their regulatory role.
\newblock \emph{Journal of biomedical informatics}, 46\penalty0 (3):\penalty0
  563--573, 2013.

\bibitem[LeCun et~al.(2015)LeCun, Bengio, and Hinton]{lecun2015deep}
LeCun, Yann, Bengio, Yoshua, and Hinton, Geoffrey.
\newblock Deep learning.
\newblock \emph{Nature}, 521\penalty0 (7553):\penalty0 436--444, 2015.

\bibitem[Lee et~al.(1993)Lee, Feinbaum, and Ambros]{lee1993c}
Lee, Rosalind~C, Feinbaum, Rhonda~L, and Ambros, Victor.
\newblock The c. elegans heterochronic gene lin-4 encodes small rnas with
  antisense complementarity to lin-14.
\newblock \emph{Cell}, 75\penalty0 (5):\penalty0 843--854, 1993.

\bibitem[Li et~al.(2015)Li, Chen, Hovy, and Jurafsky]{li2015visualizing}
Li, Jiwei, Chen, Xinlei, Hovy, Eduard, and Jurafsky, Dan.
\newblock Visualizing and understanding neural models in nlp.
\newblock \emph{arXiv preprint arXiv:1506.01066}, 2015.

\bibitem[Lorenz et~al.(2011)Lorenz, Bernhart, Zu~Siederdissen, Tafer, Flamm,
  Stadler, and Hofacker]{lorenz2011viennarna}
Lorenz, Ronny, Bernhart, Stephan~H, Zu~Siederdissen, Christian~Hoener, Tafer,
  Hakim, Flamm, Christoph, Stadler, Peter~F, and Hofacker, Ivo~L.
\newblock Viennarna package 2.0.
\newblock \emph{Algorithms for Molecular Biology}, 6\penalty0 (1):\penalty0 1,
  2011.

\bibitem[Lyngs{\o}(2004)]{lyngso2004complexity}
Lyngs{\o}, Rune~B.
\newblock Complexity of pseudoknot prediction in simple models.
\newblock In \emph{Automata, Languages and Programming}, pp.\  919--931.
  Springer, 2004.

\bibitem[Mathelier \& Carbone(2010)Mathelier and Carbone]{mathelier2010mirena}
Mathelier, Anthony and Carbone, Alessandra.
\newblock Mirena: finding micrornas with high accuracy and no learning at
  genome scale and from deep sequencing data.
\newblock \emph{Bioinformatics}, 26\penalty0 (18):\penalty0 2226--2234, 2010.

\bibitem[Tempel et~al.(2015)Tempel, Zerath, Zehraoui, Tahi,
  et~al.]{tempel2015mirboost}
Tempel, Sebastien, Zerath, Benjamin, Zehraoui, Farida, Tahi, Fariza, et~al.
\newblock mirboost: boosting support vector machines for microrna precursor
  classification.
\newblock \emph{RNA}, 21\penalty0 (5):\penalty0 775--785, 2015.

\bibitem[Xue et~al.(2005)Xue, Li, He, Liu, Li, and
  Zhang]{xue2005classification}
Xue, Chenghai, Li, Fei, He, Tao, Liu, Guo-Ping, Li, Yanda, and Zhang, Xuegong.
\newblock Classification of real and pseudo microrna precursors using local
  structure-sequence features and support vector machine.
\newblock \emph{BMC bioinformatics}, 6\penalty0 (1):\penalty0 310, 2005.

\end{thebibliography}
\bibliographystyle{mystyle}

\end{document}